\documentclass[runningheads]{llncs}
\usepackage{times}
\usepackage{soul}
\usepackage{url}
\usepackage[hidelinks]{hyperref}
\usepackage[utf8]{inputenc}
\usepackage[small]{caption}
\usepackage{graphicx}
\usepackage{amsmath}
\usepackage{booktabs}
\usepackage{algorithm}
\usepackage{algorithmic}
\usepackage{amsfonts}
\usepackage{multirow}
\usepackage{xspace}
\usepackage{xcolor}
\usepackage{enumitem}
\usepackage{hyperref}
\usepackage[misc]{ifsym}

\DeclareMathOperator{\softmax}{softmax}
\DeclareMathOperator{\entropy}{Entropy}

\newcommand{\etal}{\textit{et al.}}
\newcommand{\LSTMGANPRE}{\textsc{ATTEND-GAN$_{-SA}$}\xspace}
\newcommand{\LSTMGANNOCRIT}{\textsc{ATTEND-GAN$_{-A}$}\xspace}
\newcommand{\LSTMGAN}{\textsc{ATTEND-GAN}\xspace}
\newcommand{\CNNRNN}{\textsc{CNN+RNN}\xspace}
\newcommand{\ANPReplace}{\textsc{ANP-Replace}\xspace}
\newcommand{\ANPSCORE}{\textsc{ANP-Scoring}\xspace}
\newcommand{\RNNTRANSFER}{\textsc{RNN-Transfer}\xspace}
\newcommand{\SentiCap}{\textsc{SentiCap}\xspace}
\newcommand{\SFLSTM}{\textsc{SF-LSTM+Adap}\xspace}

\begin{document}
\title{Towards Generating Stylized Image Captions via Adversarial Training}
\author{Omid Mohamad Nezami\inst{1,2} (\Letter)
\and
Mark Dras\inst{1} \and
Stephen Wan\inst{2} \and C\'ecile Paris\inst{1,2} \and Len Hamey\inst{1}}
\authorrunning{Nezami et al.}
%
\institute{Macquarie University,
Sydney, NSW, Australia\\
\email{omid.mohamad-nezami@hdr.mq.edu.au}\\
\email{\{mark.dras,len.hamey\}@mq.edu.au}\\
\and
CSIRO's Data61, Sydney, NSW, Australia\\
\email{\{stephen.wan,cecile.paris\}@data61.csiro.au}}  

%
%
%
%
\maketitle              
\begin{abstract}
While most image captioning aims to generate objective descriptions of images, the last few years have seen work on generating visually grounded image captions which have a specific style (e.g., incorporating positive or negative sentiment). However, because the stylistic component is typically the last part of training, current models usually pay more attention to the style at the expense of accurate content description. In addition, there is a lack of variability in terms of the stylistic aspects. To address these issues, we propose an image captioning model called \LSTMGAN which has two core components: first, an attention-based caption generator to strongly correlate different parts of an image with different parts of a caption; and second, an adversarial training mechanism to assist the caption generator to add diverse stylistic components to the generated captions. Because of these components, \LSTMGAN can generate correlated captions as well as more human-like variability of stylistic patterns. Our system outperforms the state-of-the-art as well as a collection of our baseline models. A linguistic analysis of the generated captions demonstrates that captions generated using \LSTMGAN have a wider range of stylistic adjectives and adjective-noun pairs.

\keywords{Image Captioning \and Attention Mechanism \and Adversarial Training.}
\end{abstract}
\section{Introduction}
\label{sec:intro}

\begin{figure}
\begin{center}
\includegraphics[width=0.65\linewidth]{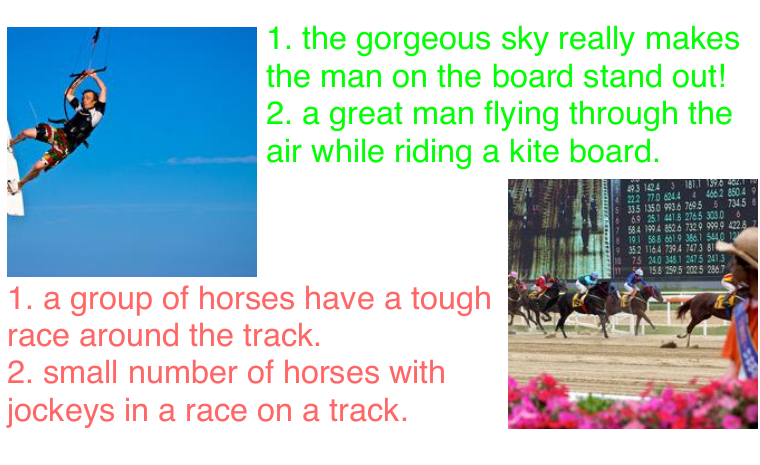}
\end{center}
\caption{Examples of positive (\textcolor{green}{green}) and negative (\textcolor{red}{red}) captions.}
\label{figure:inro}
\end{figure}

Deep learning has facilitated the task of supplying images with captions. Current image captioning models
\cite{anderson2018bottom,vinyals2015show,xu2015show} have gained considerable success due to powerful deep learning architectures and large image-caption datasets including the MSCOCO dataset \cite{lin2014microsoft}. These models mostly aim to describe an image in a factual way. Humans, however, describe an image in a way that combines subjective and stylistic properties, such as positive and negative sentiment, as in the captions of Fig.~\ref{figure:inro}. Users often find such captions more expressive and more attractive \cite{gan2017stylenet}; they have the practical purpose of enhancing the engagement level of users in social applications (e.g., chatbots) \cite{li2016share}, and can assist people to make interesting image captions in social media content \cite{gan2017stylenet}. Moreover, Mathews \etal~\cite{mathews2016senticap} found that they are more common in the descriptions of online images, and can have a role in transferring visual content clearly \cite{mathews2018semstyle}.

In stylistically enhanced descriptions, the content of images should still be reflected correctly. Moreover, the descriptions should fluently include stylistic words or phrases. To meet these criteria, previous models have used two-stage training: first, training on a large factual dataset to describe the content of an image; and then training on a small stylistic dataset to apply stylistic properties to a caption. The models have different strategies for integrating the learned information from the datasets. SentiCap has two Long Short-Term Memory (LSTM) networks: one learns from a factual dataset and the other one learns from a stylistic dataset \cite{mathews2016senticap}. 
In comparison, Gan \etal~\cite{gan2017stylenet} proposed a new type of LSTM network, factored LSTM, to learn both factual and stylistic information. The factored LSTM has three matrices instead of one multiplied to the input caption: two matrices are learned to preserve the factual aspect of the input caption and one is learned to transfer the style aspect of the input caption. 
Chen \etal~\cite{chen2018factual} applied an attention-based model which is similar to the factored LSTM, but it has an attention mechanism to differentiate attending to the factual and sentiment information of the input caption.

However, since the stylistic dataset is usually small, preserving the correlations between images and captions as well as generating a wide variety of stylistic patterns is very difficult.  An imperfect caption from the system of Mathews \etal~\cite{mathews2016senticap} --- ``a dead man doing a clever trick on a skateboard at a skate park'' --- illustrates the problem: the man is not actually dead; this is just a frequently used negative adjective.

Recently, Mathews \etal~\cite{mathews2018semstyle} dealt with this by applying a large stylistic dataset to separate the semantic and stylistic aspects of the generated captions. However, evaluation in this work was more difficult because the dataset includes stylistic captions which are not aligned to images.
To address this challenge without any large stylistic dataset, we propose \LSTMGAN, an image captioning model using an attention mechanism and a Generative Adversarial Network (GAN); our particular goal is to better apply stylistic information in the sort of two-stage architecture in previous work. Similar to this previous work, we first train a caption generator on a large factual dataset, although \LSTMGAN uses an attention-based version attending to different image regions in the caption generation process \cite{anderson2018bottom}. Because of this, each word of a generated caption is conditioned upon a relevant fine-grained region of the corresponding image, ensuring a direct correlation between the caption and the image. Then we train a caption discriminator to distinguish between captions generated by our caption generator, and real captions, generated by humans. In the next step, on a small stylistic dataset, we implement an adversarial training mechanism to guide the generator to generate sentiment-bearing captions. To do so, the generator is trained to fool the discriminator by generating correlated and highly diversified captions similar to human-generated ones. The discriminator also periodically improves itself to further challenge the generator. Because GANs are originally designed to face continuous data distributions not discrete ones like texts \cite{goodfellow2014generative}, we use a gradient policy \cite{yu2017seqgan} to guide our caption generator using the rewards received from our caption discriminator for the next generated word, as in reinforcement learning \cite{silver2016mastering}. The contributions of this paper are~\footnote{Our code and trained model are publicly available from \url{https://github.com/omidmnezami/Style-GAN}}:

\begin{itemize}[noitemsep,leftmargin=*]

    \item To generate human-like stylistic captions in a two-stage architecture, we propose \LSTMGAN (Section \ref{sec:model}) using both the designed attention-based caption generator and the adversarial training mechanism \cite{goodfellow2014generative}.
    
    \item \LSTMGAN achieves results which are significantly better than the state-of-the-art (Section \ref{sec:SotA}) and a comprehensive range of our baseline models (Section \ref{sec:base}) for generating image captions with styles.
    
    \item On the SentiCap dataset \cite{mathews2016senticap}, we show how \LSTMGAN can result in stylistic captions which are strongly correlated with visual content (Section \ref{sec:caption}). \LSTMGAN also exhibits significant variety in generating adjectives and adjective-noun pairs (Section \ref{sec:qual}).
    
\end{itemize}

\section{Related work}

\subsection{Image Captioning}

The encoder-decoder framework of Vinyals \etal~\cite{vinyals2015show} where the encoder learns to encode visual content, using a Convolutional Neural Network (CNN), and the decoder learns to describe the visual content, using a long-short term memory (LSTM) network, is the basis of modern image captioning systems. Having an attention-based component has resulted in the most successful image captioning models \cite{anderson2018bottom,rennie2017self,xu2015show,you2016image}. These models use attention in either the image side or the caption side. For instance, Xu \etal~\cite{xu2015show} and Rennie \etal\cite{rennie2017self} attended to the spatial visual features of an image. In comparison, You \etal~\cite{you2016image} applied semantic attention attending to visual concepts detected in an image. Anderson \etal~\cite{anderson2018bottom} applied an attention mechanism to attend to spatial visual features and discriminate not only the visual regions but also the detected concepts in the regions \cite{anderson2018bottom}.
In addition to factual image captioning, the ability to generate stylistic image captions has recently become popular. The key published work \cite{chen2018factual,gan2017stylenet,mathews2018semstyle,mathews2016senticap} uses a two-stage architecture, although end-to-end is possible. None of the existing work uses an adversarial training mechanism; we show this, combined with attention, significantly outperforms the previous work. 

\subsection{Generative Adversarial Network}

Goodfellow \etal~\cite{goodfellow2014generative} introduced Generative Adversarial Networks (GANs), whose training mechanism consists of a generator and a discriminator; they have been applied with great success in different applications \cite{isola2017image,liang2017recurrent,radford2015unsupervised,wang2018sentigan,yu2017seqgan}. The discriminator is trained to recognize real and synthesized samples generated by the generator. In contrast, the generator wants to generate realistic data to mislead the discriminator in distinguishing the source of data.


GANs were originally established for a continuous data space \cite{goodfellow2014generative,yu2017seqgan} rather than a discrete data distribution as in our work. To handle this, a form of reinforcement learning is usually applied, where the sentence generation process is formulated as a reinforcement learning problem \cite{silver2016mastering}; the discriminator provides a reward for the next action (in our context the next generated word), and the generator uses the reward to calculate gradients and update its parameters, as proposed in Yu \etal \cite{yu2017seqgan}.  Wang and Wan \cite{wang2018sentigan} applied this to generating sentiment-bearing text (although not conditioned on any input, such as the images in our captioning task). 

\section{\LSTMGAN Model}
\label{sec:model}
\begin{figure}
\begin{center}
\includegraphics[width=0.85\linewidth]{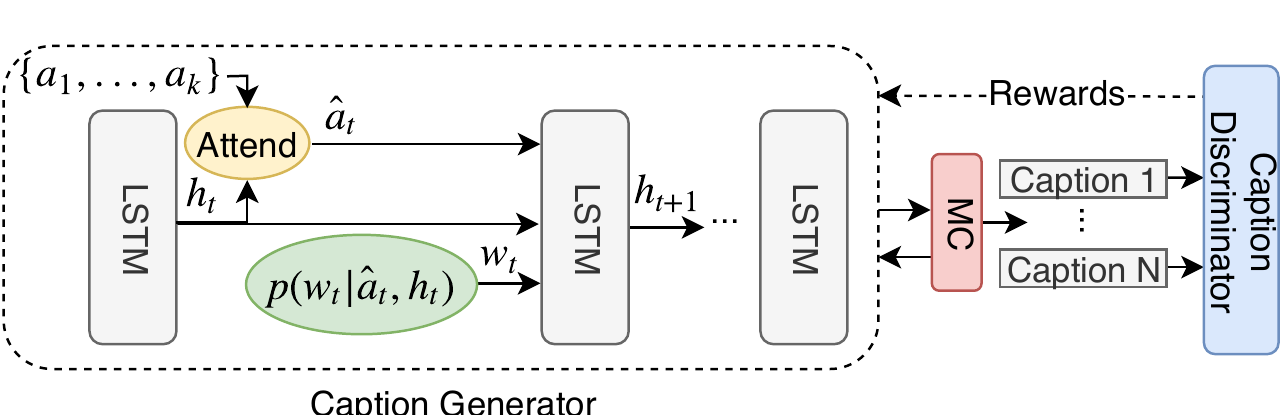}
\end{center}
\caption{The architecture of the \LSTMGAN model. $\{a_{1},...,a_{K}\}$ are spatial visual features generated by ResNet-152 network. Attend and MC modules are our attention mechanism and Monte Carlo search, respectively.}
\label{figure:attend-gan}
\end{figure}

The purpose of our image captioning model is to generate sentiment-bearing captions. Our caption generator employs an attention mechanism, described in Section \ref{sec:caption_generator}, to attend to fine-grained image regions $a=\{a_{1},...,a_{K}\}, a_{i} \in \mathbb{R}^D$, where the number of regions is $K$ with $D$ dimensions, in different time steps so as to generate an image caption $x = \{ x_1, \ldots, x_T \}, x_i \in \mathbb{R}^N$, where the size of our vocabulary is $N$ and the length of the generated caption is $T$. We also propose a caption discriminator, explained in Section \ref{sec:caption_discriminator}, to distinguish between the generated captions and human-produced ones. We describe our training in Section \ref{sec:train}. Our proposed model is called \LSTMGAN (Fig. \ref{figure:attend-gan}).

\subsection{Caption Generator}
\label{sec:caption_generator}
The goal of our caption generator $G_{\theta}(x_t|x_{1:t-1},\hat{a}_t)$ is to generate an image caption to achieve a maximum reward value from our caption discriminator $D_{\phi}(x_{1:T})$, where $\theta$ and $\phi$ are the parameters of the generator and the discriminator, respectively. 
The objective function of the generator, which is dependent on the discriminator, is to minimize:
\begin{equation}
L_{1}(\theta) = \sum_{1 \leq t \leq T}G_{\theta}(x_t|x_{1:t-1},\hat{a}_t).Z_{D_{\phi}}^{G_{\theta}}(x_{1:t})
\label{equation:gen_obj}
\end{equation}

\noindent
where $Z_{D_{\phi}}^{G_{\theta}}(x_{1:t})$ is the reward value of the partially generated sequence, $x_{1:t}$, and is estimated using the discriminator. The reward value can be interpreted as a score value that $x_{1:t}$ is real. Since the discriminator can only generate a reward value for a complete sequence, Monte Carlo (MC) search is applied, which uses the generator to roll out the remaining part of the sequence at each time step. We apply MC search $N$ times, and calculate the average reward (to decrease the variance of the next generated words):
\begin{equation}
Z_{D_{\phi}}^{G_{\theta}}(x_{1:t}) = 
\begin{cases}
\frac{1}{N} \sum\limits_{n=1}^{N} D_{\phi}(x_{1:T}^{n}), \; x_{1:T}^{n} \in MC_{G_\theta}(x_{1:t;N}) & \text{if } t<T
\\
D_{\phi}(x_{1:t}) & \text{if } t=T
\end{cases}
\label{equation:MC_search}
\end{equation}

\noindent
$x_{1:T}^{n}$ is the $n$-th MC-completed sequence at current time step $t$. In addition to Eq (\ref{equation:gen_obj}), we calculate the maximum likelihood estimation (MLE) of the generated word with respect to the attention-based content ($\hat{a}_t$) and the hidden state ($h_t$) at the current time of our LSTM, which is the core of our caption generator, as the second objective function:
\begin{equation}
L_2(\theta) = -\sum_{1 \leq t \leq T} \log( p_w(x_t \, | \, \hat{a}_t, h_t) ) + \lambda_{1} \sum_{1 \leq k \leq K}(1-\sum_{1 \leq t \leq T}a_{tk})^2
\label{equation:MLE}
\end{equation}

\noindent
$p_w$ is calculated using a multilayer perceptron with a softmax layer on its output and indicates the probabilities of the possible generated words:
\begin{equation}
p_w(x_t \, | \, \hat{a}_t, h_t) = {\softmax(\hat{a}_{t}W_a + h_{t}W_h + b_w)}
\label{equation:MLE_prob}
\end{equation}

\noindent
$W_x$ and $b_w$ are the learned weights and biases. The last term in Eq (\ref{equation:MLE}) is to encourage
our caption generator to equally consider diverse regions of the given image at the end of the caption generation process. $\lambda_1$ is a regularization parameter. $h_t$ is calculated using our LSTM:
\begin{equation}
\begin{split}
& i_t = \sigma(H_{i}h_{t-1} + W_{i}w_{t-1}+ A_{i}\hat{a}_{t} +  b_i) \\
& f_t = \sigma(H_{f}h_{t-1} + W_{f}w_{t-1} + A_{f}\hat{a}_{t} +  b_f) \\
& g_t = \tanh(H_{g}h_{t-1} + W_{g}w_{t-1} + A_{g}\hat{a}_{t} + b_g) \\
& o_t = \sigma(H_{o}h_{t-1} + W_{o}w_{t-1} + A_{o}\hat{a}_{t} + b_o) \\
& c_t = f_{t}c_{t-1}+i_{t}g_t \\
& h_t = o_t\tanh(c_t) \quad
\end{split}
\label{equation:lstm}
\end{equation}

\noindent
Here, $i_t$, $f_t$, $g_t$, $o_t$, and $c_t$ are the LSTM's gates and represent input, forget, modulation, output, and memory gates, respectively. $w_{t-1}$ is the embedded previous word in $M$ dimensions, $w_{x} \in \mathbb{R}^M$. $H_x, W_x, A_x$, and $b_x$ are learned weights and biases; and $\sigma$ is the Sigmoid function. Using $h_t$, our soft attention module generates unnormalized weights $e_{j,t}$ for each image region $a_j$. Then, the weights are normalized using a softmax layer, $e_{t}^{\prime}$:
\begin{equation}
e_{j,t} = W_e^{T} \tanh(W_{a}^{\prime}a_j+W_{h}^{\prime} h_t) ,
e_{t}^{\prime} = \softmax(e_t)
\label{equation:attention}
\end{equation}

\noindent
$W_e^{T}$ and $W_x^{\prime}$ are our trained weights. Finally, $\hat{a}_t$, our attention-based content, is calculated using Eq (\ref{equation:attention_comb}):
\begin{equation}
\hat{a}_t = \sum_{1 \leq j \leq K} e_{j,t}^{\prime}a_j
\label{equation:attention_comb}
\end{equation}

During the adversarial training, the objective function of the caption generator is a combination of Eq (\ref{equation:gen_obj}) and Eq (\ref{equation:MLE}):
\begin{equation}
L_G(\theta) = \lambda_{2} L_1(\theta) + L_2(\theta)
\label{equation:joint_loss_train}
\end{equation}

\noindent
$\lambda_2$ is a balance parameter. The discriminator cannot be learned effectively from a random initialization of the generator; we therefore pretrain the generator with the MLE objective function:

\begin{equation}
L_G(\theta) = L_2(\theta)
\label{equation:joint_loss_pretrain}
\end{equation}

\begin{algorithm}[tb]
\caption{\LSTMGAN Training Mechanism.}

\begin{algorithmic}[1]
\STATE Pre-train the caption generator ($G_{\theta}$) using Eq (\ref{equation:joint_loss_pretrain}). \\
\STATE Use $G_{\theta}$ to generate sample captions $\mathbb{P}_{G}$ and select ground-truth captions $\mathbb{P}_{H}$. \\ 
\STATE Pre-train the caption discriminator ($D_{\phi}$) using Eq (\ref{equation:WGAN_loss_dis}) and the combination of $\mathbb{P}_{G}$ and $\mathbb{P}_{H}$. \\
\REPEAT
        \FOR{$g$ steps}
            \STATE  Apply $G_{\theta}$ to generate image captions.
            \STATE  Calculate $Z_{D_{\phi}}^{G_{\theta}}$ using Eq (\ref{equation:MC_search}). \\
            \STATE  Update $\theta$, the parameters of $G_{\theta}$, using Eq (\ref{equation:joint_loss_train}).
        \ENDFOR
        \FOR{$d$ steps}
            \STATE  Generate sample captions $\mathbb{P}_{G}$ by $G_{\theta}$ and select human-generated captions $\mathbb{P}_{H}$.
            \STATE Update $\phi$, the parameters of $D_{\phi}$, using Eq (\ref{equation:WGAN_loss_dis}).
        \ENDFOR
\UNTIL{\LSTMGAN converges}
\end{algorithmic}

\label{alg:one}
\end{algorithm}

\subsection{Caption Discriminator}
\label{sec:caption_discriminator}
Our caption discriminator is inspired by the Wasserstein GAN (WGAN) \cite{arjovsky2017wasserstein} which is an improved version of the GAN \cite{goodfellow2014generative}. The WGAN generates continuous values and solves the problem of the GAN generating non-continuous outputs leading to some training difficulties (e.g. vanishing gradients). The objective function of our WGAN is:
\begin{equation}
L_D(\phi) = \mathbb{E}_{x \sim \mathbb{P}_{H}}[ D_{\phi}(x) ] -\mathbb{E}_{\overline{\rm x} \sim \mathbb{P}_{G}}[ D_{\phi}(\overline{\rm x}) ]
\label{equation:WGAN_loss_dis}
\end{equation}

\noindent
where $\phi$ are the parameters of the discriminator ($D_{\phi}$); $\mathbb{P}_{H}$ is the set of the generated captions by humans; and $\mathbb{P}_{G}$ is the set of the generated captions by the generator. $D_{\phi}$ is implemented via a Convolutional Neural Network (CNN) that calculates the score value of the input caption. To feed a caption to our CNN model, we first embed all words in the caption into $M$ embedding dimensions, $\{w^{\prime}_1, \dots, w^{\prime}_T\}, w^{\prime}_i \in \mathbb{R}^M$, and build a 2-dimensional matrix for the caption, $S \in \mathbb{R}^{T \times M}$ \cite{yu2017seqgan}. Our CNN model includes Convolutional (Conv.) layers with $P$ different kernel sizes $\{k_1,\dots, k_P\}, k_i \in \mathbb{R}^{C \times M}$, where $C$ indicates the number of the words ($C \in [1,T]$). Applying each Conv. layer to $S$ results a number of feature maps, $v_{ij} = k_i \otimes S_{j:j+C-1} + b_j$, where $\otimes$ is a convolution operation and $b_j$ is a bias vector. We apply a batch normalization layer \cite{ioffe2015batch}, and a nonlinearity, a rectified linear unit (ReLU), respectively. Then, we apply a max-pooling layer, $v_i^\ast = \max v_{ij}$. Finally, a fully connected layer is applied to output the score value of the caption. The weights of our CNN model are clipped to be in a compact space.

\subsection{\LSTMGAN Training}
\label{sec:train}
As shown in Algorithm \ref{alg:one}, we first pre-train our caption generator for a specific number of epochs. Then, we apply the best generator model to generate sample captions. The real captions are selected from the ground truth. In Step 3, our caption discriminator is pre-trained using a combination of the generated and real captions for a specific number of epochs. Here, both the caption generator and discriminator are pre-trained on a factual dataset. In Step 4, we start our adversarial training on a sentiment-bearing dataset with positive or negative sentiment. We continue the training of the caption generator and discriminator for $g$-steps and $d$-steps, respectively. Using this mechanism, we improve both the caption generator and discriminator. Here, the caption generator applies the received rewards from the caption discriminator to update its parameters using Eq (\ref{equation:joint_loss_train}).

\section{Experiments}
\subsection{Datasets}

\paragraph{Microsoft COCO Dataset.}
We use the MSCOCO image-caption dataset \cite{lin2014microsoft} to train our models. Specifically, we use the training set of the dataset including 82K+ images and 413K+ captions.

\paragraph{SentiCap Dataset.}
To add sentiment to the generated captions, our models are trained on the SentiCap dataset \cite{mathews2016senticap} including sentiment-bearing image captions. The dataset has two separate sections of sentiments: \textit{positive} and \textit{negative}. 2,873 captions paired with 998 images (409 captions with 174 images are for validation) are for training and 2019 captions paired with 673 images are for testing in the positive section. 2,468 captions paired with 997 images (429 captions with 174 images are for validation) are for training and 1,509 captions paired with 503 images are for testing in the negative section. We use the same training/test folds as in the previous work \cite{chen2018factual,mathews2016senticap}.

\subsection{Evaluation Metrics}
\label{sec:eval}

\LSTMGAN is evaluated using standard image captioning metrics: METEOR~\cite{denkowski2014meteor}, BLEU~\cite{papineni2002bleu}, CIDEr~\cite{vedantam2015cider} and ROUGE-L~\cite{lin2004rouge}. SPICE has not previously been used in the literature; however, it is reported for future comparisons because it has shown a close correlation with human-based evaluations~\cite{anderson2016spice}. Larger values of these metrics indicated better results.

\subsection{Models for Comparison}
We first trained our models on the MSCOCO dataset to generate factual captions. Then, we trained our models on the SentiCap dataset to add sentiment properties to the generated captions. This two-stage training mechanism is similar to the training methods of \cite{mathews2016senticap} and \cite{gan2017stylenet}. The work of \cite{chen2018factual}, the newest one in this domain, was also implemented in a similar way. Following this training approach makes our results directly comparable to the previous ones. Our models are compared with a range of baseline models from Mathews \etal \cite{mathews2016senticap}: \CNNRNN, which is only trained using the MSCOCO dataset; \ANPReplace, which adds the most common adjectives to a randomly chosen noun; \ANPSCORE, which applies multi-class logistic regression to select an adjective for the chosen noun; \RNNTRANSFER, which is \CNNRNN fine-tuned on the SentiCap dataset; and their key system \SentiCap, which uses two LSTM modules to learn from factual and sentiment-bearing caption. We also compare with \SFLSTM, which applies an attention mechanism to weight factual and sentiment-based information \cite{chen2018factual}. The results of all these models in Table \ref{table:result} are obtained from the corresponding references.
Moreover, we first train our attention-based model only on the factual dataset MSCOCO (we name this model \LSTMGANPRE). Second, we train our model additionally on the SentiCap dataset but without our caption discriminator (\LSTMGANNOCRIT). Finally, we train our full model using the caption discriminator (\LSTMGAN).


\subsection{Implementation Details}
\label{sec:im_details}

\paragraph{Encoder}
In this work, we apply ResNet-152~\cite{he2016deep} as our visual encoder model pre-trained using the ImageNet dataset \cite{deng2009imagenet}. In comparison with other CNN models, ResNet-152 has shown more effective results on different image-caption datasets \cite{chen2017sca}. We specifically use its Res5c layer to extract the spatial features of an image. The layer gives us $7 \times 7 \times 2048$ feature map converted to $49 \times 2048$ representing 49 semantic-based regions with 2048 dimensions. 

\paragraph{Vocabulary}
Our vocabulary has 9703 words, coming form both the MSCOCO and SentiCap datasets, for all our models. Each word is embedded into a 300 dimensional vector. 

\paragraph{Generator and Discriminator}
The size of the hidden state and the memory cell of our LSTM is set to 512. For the caption generator, we use the Adam function \cite{kingma2014adam} for optimization and set the learning rate to $0.0001$. We set the the size of our mini-batches to 64. To optimize the caption discriminator, we use the RMSprop solver \cite{tieleman2012lecture} and clip the weights to $[-0.01, 0.01]$. The mini-batches are fixed to 80 for the discriminator. We apply Monte Carlo search 5 times (Eq (\ref{equation:MC_search})). We set $\lambda_1$ and $\lambda_2$ to 1.0 and 0.1 in Eq (3) and (8), respectively. During the adversarial training, we alternate between Eq (8) and Eq (10) to optimize the generator and the discriminator, respectively. We particularly operate a single gradient decent phase on the generator ($g$ steps) and 3 gradient phases ($d$ steps) on the discriminator every time. The models are trained for 20 epochs to converge. The METEOR metric is used to select the model with the best performance on the validation sets of positive and negative datasets of SentiCap because it has a close correlation with human judgments and is less computationally expensive than SPICE which requires dependency parsing \cite{anderson2016spice}.

\subsection{Results: Comparison with the State-of-the-art}
\label{sec:SotA}
All models in Table~\ref{table:result} used the same training/test folds of the SentiCap dataset to make them comparable. In comparison with the state-of-the-art, our full model (\LSTMGAN) achieves the best results for all image captioning metrics in both positive and negative parts of the SentiCap dataset. We report the average results to show the average improvements of our models over the state-of-the-art model. \LSTMGAN achieved large gains of $6.15$, $6.45$, $3.00$, and $2.95$ points with respect to the best previous model using BLEU-1, ROUGE-L, CIDEr and BLEU-2 metrics, respectively. Other metrics show smaller but still positive improvements.


\subsection{Results: Comparison with our Baseline Models}
\label{sec:base}
Our models are compared in Table \ref{table:result} in terms of image captioning metrics. \LSTMGAN outperforms \LSTMGANNOCRIT over all metrics across both positive and negative parts of the SentiCap dataset; the discriminator is thus an important part of the architecture.
\LSTMGAN outperforms \LSTMGANPRE for all metrics except, by a small margin, CIDEr and ROUGE-L.  Recall that \LSTMGANPRE is trained only on the large MSCOCO (with many captions), and so is in a sense encouraged to have diverse captions; second-stage training for \LSTMGANNOCRIT and \LSTMGAN leads to more focussed captions relevant to SentiCap.  As CIDEr and ROUGE-L are the two recall-oriented metrics, they suffer in this two-stage process, illustrating the issue we noted in Sec~\ref{sec:intro}. The discriminator, however, removes almost all of this penalty, as well as boosting the other metrics beyond \LSTMGANPRE. Furthermore, Sec~\ref{sec:qual} illustrates how \LSTMGANPRE produces unsatisfactory captions in terms of sentiment.

\begin{table}[t!]
\centering
\caption{The compared performances on different sections of SentiCap and their average. BLEU-N metric is shown by B-N. (The best results are bold.)}
\begin{tabular}{ |l |l |c |c |c |c |c |c |c |c| }
\hline
Senti & Model & B-1 & B-2 & B-3 & B-4 &  ROUGE-L & METEOR & CIDEr & SPICE \\
\hline
\multirow{10}{*}{Pos}
& CNN+RNN & 48.7 & 28.1 & 17.0 & 10.7 & 36.6 & 15.3 & 55.6 & \_ \\
& ANP-Replace & 48.2 & 27.8 & 16.4 & 10.1 & 36.6 & 16.5 & 55.2 & \_ \\
& ANP-Scoring & 48.3 & 27.9 & 16.6 & 10.1 & 36.5 & 16.6 & 55.4 & \_ \\
& RNN-Transfer & 49.3 & 29.5 & 17.9 & 10.9 & 37.2 & 17.0 & 54.1 & \_ \\
& SentiCap & 49.1 & 29.1 & 17.5 & 10.8 & 36.5 & 16.8 & 54.4 & \_ \\
& SF-LSTM + Adap & 50.5 & 30.8 & 19.1 & 12.1 & 38.0 & 16.6 & 60.0 & \_ \\ 
& Ours: \LSTMGANPRE  & 56.1 & 32.5 & 19.4 & 11.8 & 44.8 & 17.1 & 63.0 & 15.9 \\
& Ours: \LSTMGANNOCRIT & 55.8 & 33.4 & 20.1 & 12.4 & 44.2 & 18.6 & 61.1 & 15.7  \\
& Ours: \LSTMGAN & 56.9 & 33.6 & 20.3 & 12.5 & 44.3 & 18.8 & 61.6 & 15.9 \\
\hline
\multirow{10}{*}{Neg}
& CNN+RNN & 47.6 & 27.5 & 16.3 & 9.8 & 36.1 & 15.0 & 54.6 & \_ \\
& ANP-Replace & 48.1 & 28.8 & 17.7 & 10.9 & 36.3 & 16.0 & 56.5 & \_ \\
& ANP-Scoring & 47.9 & 28.7 & 17.7 & 11.1 & 36.2 & 16.0 & 57.1 & \_ \\
& RNN-Transfer & 47.8 & 29.0 & 18.7 & 12.1 & 36.7 & 16.2 & 55.9 & \_ \\
& SentiCap & 50.0 & 31.2 & 20.3 & 13.1 & 37.9 & 16.8 & 61.8 & \_ \\
& SF-LSTM + Adap & 50.3 & 31.0 & 20.1 & 13.3 & 38.0 & 16.2 & 59.7 & \_ \\
& Ours: \LSTMGANPRE  & 55.4 & 32.4 & 19.4 & 11.9 & 44.4 & 17.0 & 63.4 & 15.6 \\
& Ours: \LSTMGANNOCRIT  & 54.7 & 32.6 & 20.4 & 12.9 & 43.2 & 17.7 & 60.4 & 16.1 \\
& Ours: \LSTMGAN  & 56.2 & 34.1 & 21.3 & 13.6 & 44.6 & 17.9 &  64.1 & 16.2 \\
\hline
\multirow{10}{*}{Avg}
& CNN+RNN & 48.15 & 27.80 & 16.65 & 10.25 & 36.35 & 15.15 & 55.10 & \_ \\
& ANP-Replace & 48.15 & 28.30 & 17.05 & 10.50 & 36.45 & 16.25 & 55.85 & \_ \\
& ANP-Scoring & 48.10 & 28.30 & 17.15 & 10.60 & 36.35 & 16.30 & 56.25 & \_  \\
& RNN-Transfer & 48.55 & 29.25 & 18.30 & 11.50 & 36.95 & 16.60 & 55.00 & \_  \\
& SentiCap & 49.55 & 30.15 & 18.90 & 11.95 & 37.20 & 16.80 & 58.10 & \_ \\
& SF-LSTM + Adap & 50.40 & 30.90 & 19.60 & 12.70 & 38.00 & 16.40 & 59.85 & \_ \\
& Ours: \LSTMGANPRE  & 55.75 & 32.45 & 19.40 & 11.85 & \textbf{44.60} & 17.05 & \textbf{63.20} & 15.75 \\
& Ours: \LSTMGANNOCRIT  & 55.25 & 33.00 & 20.25 & 12.65 & 43.70 & 18.15 & 60.75 & 15.90 \\
& Ours: \LSTMGAN & \textbf{56.55} & \textbf{33.85} & \textbf{20.80} & \textbf{13.05} & 44.45 & \textbf{18.35} & 62.85 & \textbf{16.05} \\
\hline
\end{tabular}
\label{table:result}
\end{table}



\begin{table}
\centering
\caption{Entropy and $Top_4$ of the generated adjectives using different models.}
\begin{tabular}{ |l|l|l|l| }
\hline
Senti & Model & Entropy & Top$_4$ \\ 
\hline
\multirow{3}{*}{Pos}
& \LSTMGANPRE & 2.2457 & 93.33\% \\
& \LSTMGANNOCRIT & 3.0324 &  72.11\%  \\
& \LSTMGAN & 3.5671 &  62.33\%   \\
 \hline
\multirow{3}{*}{Neg}
& \LSTMGANPRE & 2.2448 & 91.67\%  \\
& \LSTMGANNOCRIT & 4.1040  & 48.44\%  \\
& \LSTMGAN & 3.9562 & 50.51\%   \\
\hline
\multirow{3}{*}{Avg}
& \LSTMGANPRE & 2.2453  & 92.50\%   \\
& \LSTMGANNOCRIT & 3.5682 & 60.28\%  \\
& \LSTMGAN & \textbf{3.7617}  & \textbf{56.42\%}   \\
\hline
\end{tabular}
\label{table:result_entropy}
\end{table}

\begin{table}
\centering
\caption{The top-10 adjectives that are generated by our models and are in the adjective-noun pairs of the SentiCap dataset.}
\begin{tabular}{ |l |l |l| }
\hline
Senti & Model & Top 10 Adjectives \\ 
\hline
\multirow{3}{*}{Pos}
& \LSTMGANPRE & white, black, small, blue, different, little, busy, \_, \_, \_ \\
& \LSTMGANNOCRIT & nice, beautiful, happy, busy, great, sunny, good, cute, pretty, white   \\
& \LSTMGAN & nice, beautiful, happy, great, good, sunny, busy, white, pretty, delicious   \\
 \hline
\multirow{3}{*}{Neg}
& \LSTMGANPRE  & black, white, small, blue, different, tall, little, \_, \_ , \_   \\
& \LSTMGANNOCRIT & lonely, dead, broken, stupid, dirty, bad, cold, little, crazy, lazy \\
& \LSTMGAN & lonely, stupid, broken, dirty, dead, cold, bad, white, crazy, little \\
\hline
\end{tabular}
\label{table:result_adj}
\end{table}

\subsection{Qualitative Results}
\label{sec:qual}
To analyze the quality of language generated by our models, we extract all generated adjectives using the Stanford part-of-speech tagger software \cite{toutanova2003feature}, and select the adjectives found in the adjective-noun pairs (ANPs) of the SentiCap dataset. Then, we calculate Entropy of the distribution of these adjectives as a measure of variety in lexical selection (higher scores mean more variety) using Eq (\ref{equation:entropy}).
\begin{equation}
\entropy = -\sum_{1 \leq j \leq U}{\log_2[p(A_j)]} \times p(A_j) \quad
\label{equation:entropy}
\end{equation} 

\noindent
where $p(A_j)$ is the probability of the adjective ($A_j$) and $U$ indicates the number of all unique adjectives. Moreover, we calculate the total probability mass of the four most frequent adjectives (Top$_4$) generated by our models. Here, lower values mean that the model allocates more probability to other generated adjectives, also indicating greater variety.


Table \ref{table:result_entropy} shows that \LSTMGAN achieves the best results on average for Entropy (highest score) and Top$_4$ (lowest) compared to other models, by a large margin with respect to \LSTMGANPRE. It is not surprising that \LSTMGANPRE has the lowest variability of use of sentiment-bearing adjectives because it does not use the stylistic dataset. As demonstrated by the improvement of \LSTMGAN over \LSTMGANNOCRIT, the discriminator helps in generating a greater diversity of adjectives.


The top-10 adjectives generated by our models are shown in Table \ref{table:result_adj}. ``white'' is generated for both negative and positive sections because they are common in both sections. \LSTMGAN and \LSTMGANNOCRIT produce a natural ranking of sentiment-bearing adjectives for both sections. For example, these models rank ``nice'' as the most positive adjective, and ``lonely'' as the most negative. As \LSTMGANPRE does not use the stylistic dataset, it generates a similar and limited ($<10$) range of adjectives for both. 

\begin{figure}
\begin{center}
\includegraphics[width=1.0\linewidth]{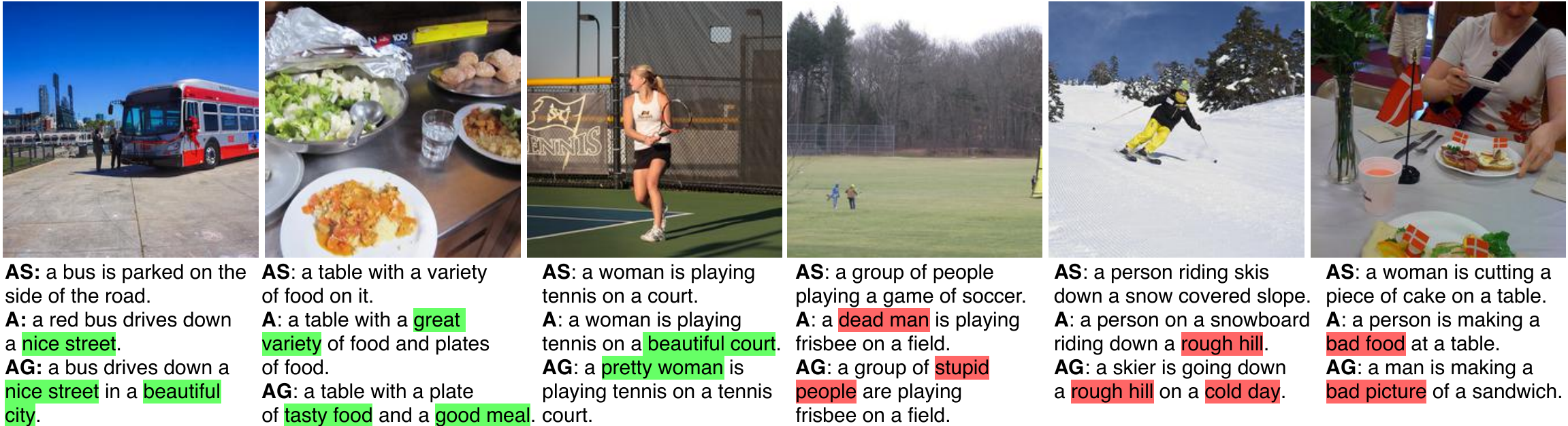}
\end{center}
\caption{Examples on the positive (first 3) and negative (last 3) datasets (AS for \LSTMGANPRE, A for \LSTMGANNOCRIT and AG for \LSTMGAN). Green and red colors indicate the generated positive and negative adjective-noun pairs in SentiCap, resepectivelly.}
\label{figure:samples}
\end{figure}

\subsection{Generated Captions}
\label{sec:caption}
Fig. \ref{figure:samples} shows sample sentiment-bearing captions generated by our models for the positive and negative sections of the SentiCap dataset.\footnote{See a link to supplementary materials for additional samples: \url{https://github.com/omidmnezami/Style-GAN/blob/master/st.pdf}.} For instance, for the first two images, \LSTMGAN correctly applies positive sentiments to describe the corresponding images (e.g., ``nice street'', ``tasty food''). Here, \LSTMGANNOCRIT also succeeds in generating captions with positive sentiments, but less well. In the third image, \LSTMGAN uses ``pretty woman'' to describe the image which is better than the ``beautiful court'' of \LSTMGANNOCRIT: for this image, all ground-truth captions have positive sentiment for the noun ``girl'' (e.g. ``a beautiful girl is running and swinging a tennis racket''); none of them describes the noun ``court'' with a sentiment-bearing adjective as \LSTMGANNOCRIT does. 
For all images, since \LSTMGANPRE is not trained using the SentiCap dataset, it does not generate any caption with sentiment. For the fourth image, \LSTMGAN generates ``a group of stupid people are playing frisbee on a field'', applying ``stupid people'' to describe the image negatively. Here, one of the ground-truth captions exactly includes ``stupid people'' (``two stupid people in open field watching yellow tent blown away'').  \LSTMGANNOCRIT, like our flawed example from Sec~\ref{sec:intro}, refers instead inaccurately to a dead man.
For the fifth image (as for the first image), \LSTMGAN has incorporates more (appropriate) sentiment in comparison to \LSTMGANNOCRIT. It generates ``rough hill'' and ``cold day'', while \LSTMGANNOCRIT only generates the former. It also uses ``skier'' which is more appropriate than ``person''. In the last image, \LSTMGAN adds ``bad picture'' and \LSTMGANNOCRIT generates ``bad food''. One of the ground-truth captions exactly includes ``bad picture''.

\section{Conclusion}
In this paper, we proposed \LSTMGAN, an attention-based image captioning model using an adversarial training mechanism. Our model is capable of generating stylistic captions which are strongly correlated with images and contain diverse stylistic components. \LSTMGAN achieves the state-of-the-art performance on the SentiCap dataset. It also outperforms our baseline models and generates stylistic captions with a high level of variety. Future work includes developing \LSTMGAN to generate a wider range of captions and developing further mechanisms to ensure compatibility with the visual content.

\bibliographystyle{splncs04}
\bibliography{ref}
%




\end{document}